\documentclass[11pt]{article}

\usepackage[margin=1in]{geometry}
\usepackage{amsmath,amssymb}
\usepackage{graphicx}
\usepackage{booktabs}
\usepackage{array}
\usepackage{natbib}
\usepackage{url}
\usepackage{microtype}
\usepackage{placeins}
\usepackage{hyperref}

\graphicspath{{figures/}}
\hypersetup{
  colorlinks=true,
  citecolor=black,
  linkcolor=black,
  urlcolor=blue,
  pdftitle={Human and LLM Facilitation in Collaborative Design Meetings: A Computational Behavioral Comparison},
  pdfauthor={Mohammed Azeez Khan, Aaron D'Souza, Ashutosh Mishra, Ummul Faiz Z. Bibi, Neha K. Nair, Arshaan Khan},
  pdfkeywords={LLM facilitation, large language models, human-AI collaboration, human-LLM comparison, meeting facilitation, collaborative design, conversational AI, behavioral analysis, design meetings, AMI Meeting Corpus}
}
\newcommand{\doi}[1]{\href{https://doi.org/#1}{doi: \nolinkurl{#1}}}

\title{Human and LLM Facilitation in Collaborative Design Meetings:\\ A Computational Behavioral Comparison}

\author{%
Mohammed Azeez Khan\textsuperscript{1},
Aaron D'Souza\textsuperscript{2},
Ashutosh Mishra\textsuperscript{2},\\
Ummul Faiz Z.\ Bibi\textsuperscript{3},
Neha K.\ Nair\textsuperscript{4},
and Arshaan Khan\textsuperscript{3}
}
\date{}

\begin{document}
\maketitle
\vspace{-1.2em}
\begin{center}
\small
\textsuperscript{1}Department of Computer Science and Engineering, National Institute of Technology Warangal, Warangal, India\\
\textsuperscript{2}Department of Electronics and Communication Engineering, National Institute of Technology Warangal, Warangal, India\\
\textsuperscript{3}Department of Computer Science and Engineering,\\
Nawab Shah Alam Khan College of Engineering and Technology, Hyderabad, India\\
\textsuperscript{4}Department of Physics, National Institute of Technology Warangal, Warangal, India
\end{center}
\vspace{0.6em}

\begin{abstract}
Large language models (LLMs) are increasingly being explored as facilitators of group discussion and collaborative work, but the behavioral differences between LLM-generated and human facilitation remain insufficiently characterized. We compare human Project Manager facilitation turns from scenario-based collaborative design meetings with Llama~3.1 8B Instruct responses across 199 paired contexts from the AMI Meeting Corpus. The analysis covers five confirmatory dimensions: contextual semantic distance, a lexical elaboration proxy, empathy-and-hedge lexical density, topic divergence, and phase-vocabulary alignment. An exploratory directiveness measure failed domain validation and is not interpreted. The sampled human Project Manager turns were higher on contextual semantic distance ($r = 0.644$) and on empathy-and-hedge lexical density ($r = 0.203$). The second contrast is carried mainly by hedge terms, so we read it as lexical hedging rather than psychological empathy. Llama-generated facilitation was higher on topic divergence ($r = 0.703$), the lexical elaboration proxy ($r = 0.288$; length-confounded), and instruction-conditioned phase-vocabulary alignment ($r = 0.382$). A secondary no-phase check left the Llama-higher phase contrast in place. The system prompt still requested a facilitation move, so that check is not evidence of spontaneous phase awareness. These differences should not be interpreted as a ranking of facilitation quality. The resulting behavioral profile provides a reference point for later comparisons of LLM facilitators, human facilitators, prompts, models, and collaborative settings.
\end{abstract}


\section{Introduction}
Large language models (LLMs) are increasingly being explored as facilitators of group discussion, collaboration, and decision making \citep{korre2025, dong2024facilitation, alsobay2026table}. Related work has also examined generative AI in design \citep{peng2025} and designer co-creation with AI tools \citep{gmeiner2023}. Work on LLM facilitation has begun to study facilitation systems and group outcomes. A related but distinct question concerns the linguistic form of facilitation itself. We compare LLM-generated and human facilitation in collaborative design meetings.

Existing research has examined how LLM-based facilitation can be implemented, when LLMs should intervene, and how facilitation affects group interaction or outcomes. A complementary question is how the linguistic behavior of LLM facilitators differs from observed human facilitation in comparable scenario-based meetings. We address this question through a context-matched computational behavioral comparison of human and LLM facilitation in collaborative design meetings.

The present analysis uses scenario meetings from the AMI Meeting Corpus \citep{carletta2005ami} and Llama~3.1 8B Instruct \citep{dubey2024llama3}. We compare LLM-generated facilitation with human Project Manager turns on five confirmatory dimensions: contextual semantic distance, a lexical elaboration proxy, empathy-and-hedge lexical density, topic divergence, and phase-vocabulary alignment. Directiveness is exploratory and unvalidated. These scores describe facilitation language.

The research questions are:
\begin{enumerate}
\item What behavioral differences distinguish LLM-generated facilitation from human facilitation in collaborative design meetings?
\item Which measurable behavioral dimensions most strongly differentiate LLM-generated and human facilitation turns?
\item How robust are these behavioral differences to decoding temperature and sensitivity to representation, response length, duplicate contexts, and phase prompting?
\end{enumerate}

The contributions are as follows.
\begin{enumerate}
\item A context-matched behavioral comparison of human and LLM facilitation using 199 paired contexts.
\item A multi-dimensional computational characterization spanning contextual semantic distance, lexical elaboration, hedge-and-empathy lexical density, topic divergence, and phase-vocabulary alignment. These measures are not presented as a validated universal taxonomy.
\item Robustness analysis across response length, duplicate contexts, LDA topic count, TF--IDF representation, decoding temperature, phase-label removal, and mixed-effects modeling.
\item A behavioral reference point for future research on LLM facilitation. It is not offered as a universal benchmark.
\end{enumerate}

\section{Literature Review}
\subsection{LLM facilitation systems}
Korre \emph{et al.} \citep{korre2025} survey how LLMs are being used to evaluate and facilitate online discussions, including intervention strategies and open research directions. Dong, Ding, and Ito \citep{dong2024facilitation} designed a multi-phase facilitation agent that uses GPT-3.5 to produce facilitation messages for large-scale online discussion. That work is a system design: it specifies when a phase-based agent should speak and how an LLM is controlled. It does not compare the wording of those messages with human facilitation in matched conversational contexts.

\subsection{Human and LLM facilitation behavior}
Tsirmpas, Korre, and Pavlopoulos \citep{tsirmpas2026facilitate} studied a different behavioral question: when a facilitator should intervene. In their survey of online-discussion excerpts, human experts were more cautious about intervening, whereas the LLMs they tested were more eager to facilitate. Their comparison concerns the decision to intervene, not the linguistic form of a facilitation turn written for the same preceding context. Neighboring studies of human--LLM co-writing \citep{lee2022} and story writing \citep{yuan2022} likewise characterize collaborative language generation, but not meeting facilitation.

\subsection{Group outcomes}
Alsobay \emph{et al.} \citep{alsobay2026table} ran a preregistered experiment in which live groups received no facilitation, a one-time prompt, a human facilitator, or a GPT-4o facilitator. LLM facilitation increased information sharing by raising the minimum level of task engagement. Neither human nor LLM facilitation significantly changed the final group decision. That result is about downstream participation and decision outcomes. It does not describe the computational properties of the facilitation language itself.

\subsection{Where the comparison fits}
Existing work asks how a facilitation system is built, when it should intervene, and what happens to the group afterward. We instead characterize the behavioral properties of the facilitation language itself. The comparison uses context-matched human and LLM facilitation turns from scenario-based collaborative design meetings. In that setting, Starostka \emph{et al.} \citep{starostka2021} distinguished a method approach to design-thinking facilitation from a cofacilitation approach, and Seidel and Fixson \citep{seidel2013} studied how novice teams use needfinding, brainstorming, and prototyping. Brown \citep{brown2008} described design work as movement through inspiration, ideation, and implementation rather than a fixed sequence of steps. Protocol studies of design communication \citep{stempfle2002, kan2017} and computational displays of spoken meetings \citep{chandrasegaran2019} show that meeting language can be measured, but they do not supply a human--LLM facilitation comparison. Yu \emph{et al.} \citep{peng2025} discuss opportunities and reliability limits of generative AI in design, including hallucination. Gmeiner \emph{et al.} \citep{gmeiner2023} studied how designers learn to co-create with AI manufacturing tools. Neither paper measures facilitation turns. Sentence embeddings \citep{reimers2019}, latent Dirichlet allocation \citep{blei2003lda}, and a surface lexical pipeline are the instruments for that comparison.

\section{Method}
\subsection{Corpus and data processing}
The analyzed meetings are elicited scenario meetings from the AMI Meeting Corpus \citep{carletta2005ami, carletta2006elra}. They are not a sample of generic workplace meetings. In the scenario, four participants (a project manager, an industrial designer, a user-interface designer, and a marketing expert) take a project through four meetings in a day \citep{carletta2006elra}. The design task is a new remote control \citep{ami2006scenario}. The corpus release also contains other meetings; this comparison uses the scenario design meetings and the project-manager turns within them.

We used the text transcription subset comprising 279 transcript files. Transcripts were segmented into sentence-level speaker turns, yielding 89,472 sentence-turns across 279 sessions. In the AMI scenario, the project manager runs the meetings \citep{carletta2006elra}, and those turns are the source of the human facilitation sample. Session phases were assigned by relative position in each transcript (first 20\% $\rightarrow$ Empathize/Define; middle 50\% $\rightarrow$ Ideate; final 30\% $\rightarrow$ Prototype/Evaluate). These labels are an author-defined temporal heuristic. They are not a published phase model, and they are not ground-truth phase annotation.

Candidate facilitation moves were extracted from Project Manager turns with rule-based filters: (a)~word count between 8 and 60 words, (b)~absence of casual non-meeting fragments, (c)~presence of facilitation signals (interrogative syntax, facilitation keywords such as ``what if'' or ``how might,'' or modal verbs in turns of at least 12 words), and (d)~co-occurrence with design-meeting vocabulary. This process identified 2,685 candidate facilitation moves across 139 sessions. Of those candidates, 152 were labeled Ideate and 2,533 Prototype/Evaluate. None fell in the first temporal window. The evaluated set contains 199 of these candidates: 149 Ideate and 50 Prototype/Evaluate. The released preprocessing script reconstructs that set from the context identifiers stored with the generated responses. It does not record an additional random draw from the candidate pool. The evaluated human sample therefore consists of rule-selected candidate facilitation moves rather than an exhaustive or independently validated annotation of PM facilitation behavior. A 60-turn sheet for facilitation-move agreement was prepared, but the ratings were left blank, so no inter-rater coefficient is available.

\subsection{LLM inference setup}
Matched LLM facilitation responses were generated using Meta Llama~3.1~8B~Instruct \citep{dubey2024llama3}. Preprocessing, feature extraction, and all statistical analyses were run locally on a laptop with an NVIDIA GeForce RTX 3050 Ti (4 GB VRAM). That GPU cannot hold Llama~3.1 8B, so response generation was the step run on Google Colab, in 4-bit format via \texttt{bitsandbytes} (NF4 quantization, double quantization, float16 compute dtype) on an NVIDIA Tesla T4 (16 GB VRAM).

For each of the 199 human contexts, a prompt was constructed containing a system prompt defining the facilitator role (a single move of 1--3 sentences) and a user prompt containing the preceding 5 conversational turns (truncated to a maximum of 180 words), the explicit design-phase label, and the generation request. Context matching refers to the same extracted conversational window supplied to the LLM, not to equal informational access between the LLM and the human PM during the original meeting. Responses were generated at three temperatures ($T \in \{0.3, 0.7, 1.0\}$) with \texttt{do\_sample=True}, \texttt{top\_p=0.9}, and \texttt{max\_new\_tokens=150}, producing 597 valid LLM responses. A response counted as valid only if it contained at least five words, contained none of the refusal phrases ``as an ai,'' ``i cannot,'' ``i'm unable,'' ``i am unable,'' ``i don't have,'' ``as a language model,'' and ``i'm just,'' and, when the context contained more than ten words, shared no more than 90\% of its word types with that context. A random seed for this primary generation is not recorded in the released generation script.

The primary LLM prompt explicitly includes the temporal phase label (for example, ``Current design phase: Ideate''). A secondary generation repeated the same 199 contexts, model, 4-bit settings, temperatures, and sampling options with that line removed (seed = 42; all 597 responses passed the existing validity filter). Phase-vocabulary alignment was rescored with the original token-proportion function. This contrast is a sensitivity analysis of the phase-label confound and is not added to the five-dimension confirmatory family. The system prompt still asks for a design-facilitation move, so a no-phase response is not an uninstructed turn.

\subsection{Behavioral dimensions and operationalization}
Each facilitation move was evaluated on six computational dimensions. Five are confirmatory. Directiveness is exploratory.

\begin{enumerate}
\item \textbf{Contextual semantic distance ($F_1$).} $F_1 = 1 - \cos(\mathbf{e}_{\text{ctx}}, \mathbf{e}_{\text{resp}})$, with embeddings from \texttt{all-MiniLM-L6-v2}. Cosine comparison of sentence embeddings follows Sentence-BERT \citep{reimers2019}. Higher values indicate greater embedding-space distance from the preceding context.

\item \textbf{Directiveness (exploratory / unvalidated).} Probability $P(\text{directive})$ from a DistilBERT classifier \citep{sanh2019distilbert} trained on 197 synthetic examples (99 directive, 98 open-ended). Human domain validation on 100 AMI utterances yielded Cohen's $\kappa = -0.0231$, indicating chance-level agreement. This measure is excluded from confirmatory inference.

\item \textbf{Lexical elaboration proxy ($F_3$).} Composite of named-entity density, type--token ratio, and normalized average word length. Named-entity density uses spaCy \citep{spacy2017}. Normalized word length is $\min(\mathrm{AWL}/10, 1)$. This metric captures surface lexical elaboration. The robustness analysis investigates word-count confounding.

\item \textbf{Empathy-and-hedge lexical density ($F_4$).} Proportion of empathy and hedging terms relative to turn word count. The score counts dictionary hits. Median scores are zero for both groups, and the results section reports which part of the dictionary carries the contrast.

\item \textbf{Topic divergence ($F_5$).} Jensen--Shannon distance between context and response topic distributions from a 15-topic LDA model \citep{blei2003lda} trained jointly on all human and LLM texts. The number of topics was selected heuristically. Sensitivity to that choice is reported below.

\item \textbf{Phase-vocabulary alignment ($F_6$).} Proportion of response tokens matching predefined phase-specific vocabulary lists. The primary score uses responses generated with the phase label in the prompt. The no-label generation is reported separately as a sensitivity analysis.
\end{enumerate}

\subsection{Statistical analysis}
\label{sec:stats}
The 199 human PM turns and 199 LLM responses at $T = 0.7$ form matched pairs by context identifier. The primary test is the paired Wilcoxon signed-rank test \citep{wilcoxon1945}. Bonferroni correction is applied once across the five confirmatory dimensions. A test is rejected when its raw $p$-value is below $\alpha/5 = 0.010$, which is the same decision as rejecting when the adjusted value $p_\text{Bonf} = \min(1, 5p)$ is below $0.05$. Comparing $p_\text{Bonf}$ with $0.010$ would apply the correction twice. Effect sizes are reported as $r = |Z|/\sqrt{n}$ from the normal approximation \citep{rosenthal1994parametric}, with 95\% bootstrap confidence intervals (2{,}000 resamples, seed = 42). Directiveness is excluded from Bonferroni correction and treated as exploratory.

Two paired Wilcoxon tests at $T = 0.7$ (human versus no-phase, and label-present versus no-phase) share a separate Bonferroni correction over those two tests. They do not change the significance status of the five confirmatory dimensions. Temperature-specific label contrasts are Bonferroni-adjusted across the three temperatures and are not added to either family as confirmatory tests.

The 597 LLM responses are 199 contexts crossed with three temperatures. The inferential temperature test is a Friedman test within context, with Bonferroni correction over the five confirmatory dimensions. A Spearman correlation on the stacked 597 rows is reported only as a description, because that correlation does not model the repeated contexts.

The robustness checks are as follows. (a)~Length-matched subsampling keeps pairs whose relative word-count difference is below 0.50, which leaves $n = 73$ pairs, and repeats the paired Wilcoxon test. The relative difference is $|\mathrm{WC}_{\mathrm{human}} - \mathrm{WC}_{\mathrm{LLM}}| / (\mathrm{WC}_{\mathrm{human}} + 1)$. (b)~A Spearman correlation relates each paired difference to human word count. (c)~A linear mixed-effects model uses an LLM indicator as the fixed effect and the context identifier as a random intercept.

\section{Results}
\label{sec:results}
\subsection{Overall comparison}
\begin{figure}[htbp]
    \centering
    \includegraphics[width=0.92\textwidth]{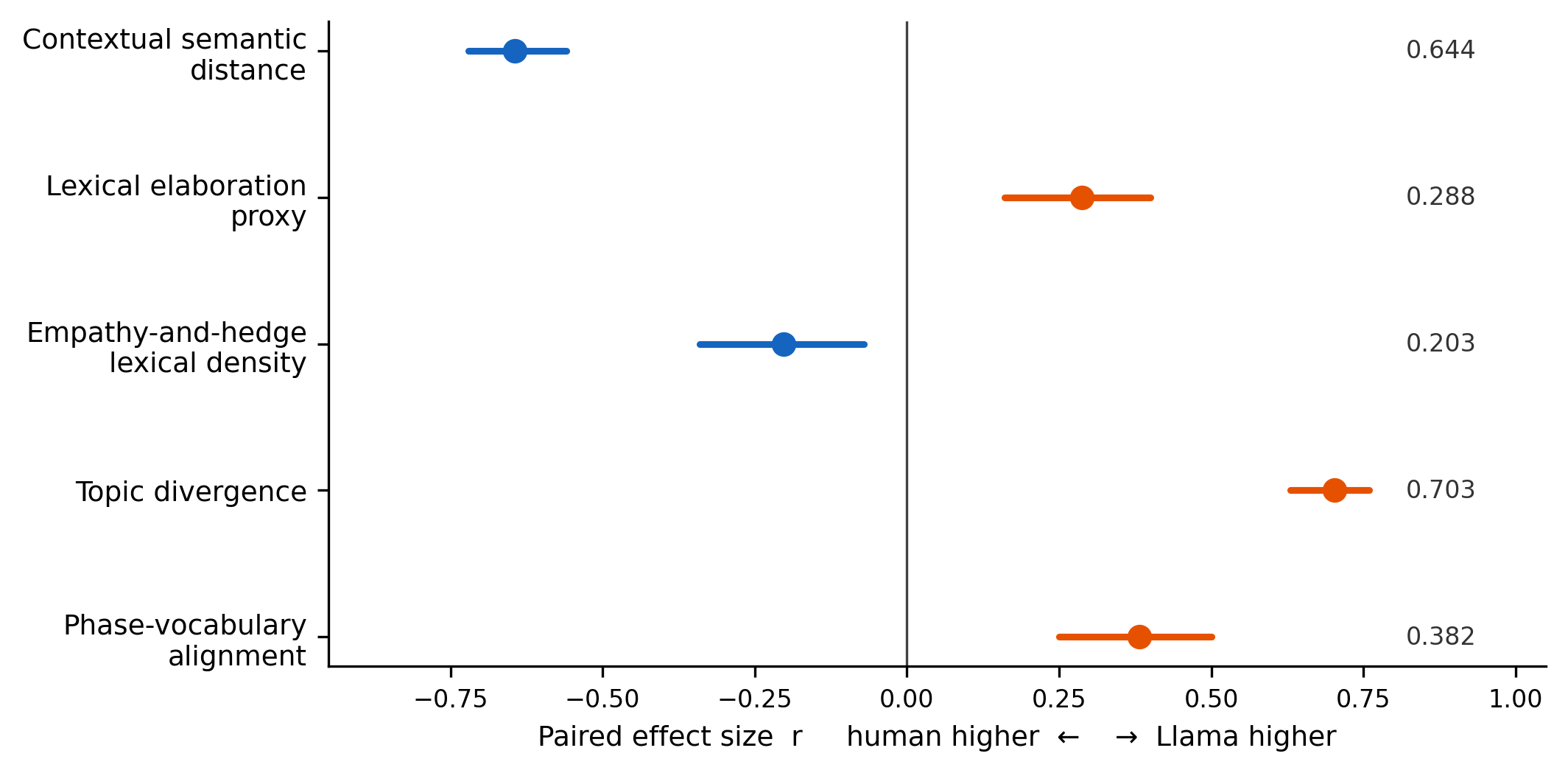}
    \caption{Primary effect sizes at $T = 0.7$ ($n = 199$ pairs). Points are the paired effect size $r$, signed so that negative values mean the human turn is higher and positive values mean the Llama response is higher. Intervals are the 95\% bootstrap confidence intervals on $r$, given the same sign. Directiveness is omitted because it is unvalidated.}
    \label{fig:forest}
\end{figure}

The sampled human Project Manager turns and the Llama responses differed across the five confirmatory computational dimensions (Figure~\ref{fig:forest}; Table~\ref{tab:main_results}). At $T = 0.7$, human Project Manager turns were higher on contextual semantic distance and on empathy-and-hedge lexical density. Llama~3.1 8B Instruct responses were higher on topic divergence, the lexical elaboration proxy, and phase-vocabulary alignment. All five contrasts met the primary Bonferroni rule: raw $p < 0.010$, equivalently $p_\text{Bonf} < 0.05$. Table~\ref{tab:bounds} states what each contrast supports and what it does not. LLM responses were longer (mean $38.5$ words at $T = 0.7$) than human turns (mean $23.5$ words; Table~\ref{tab:descriptive}), which matters for the length-sensitive scores.

\begin{table}[htbp]
\caption{What the five confirmatory contrasts support. Directions are human versus Llama~3.1 8B Instruct at $T = 0.7$.}
\label{tab:bounds}
\centering
\small
\begin{tabular}{@{}>{\raggedright\arraybackslash}p{0.20\textwidth}>{\raggedright\arraybackslash}p{0.16\textwidth}>{\raggedright\arraybackslash}p{0.28\textwidth}>{\raggedright\arraybackslash}p{0.26\textwidth}@{}}
\toprule
\textbf{Dimension} & \textbf{Observed difference} & \textbf{What it supports} & \textbf{What it does not establish} \\
\midrule
Contextual semantic distance & Human higher & Greater distance in \texttt{all-MiniLM-L6-v2} space & General semantic superiority \\
Lexical elaboration & Llama higher & Greater surface lexical elaboration & Conceptual specificity \\
Empathy/hedging & Human higher & A lexical hedge/empathy difference & Psychological empathy \\
Topic divergence & Llama higher & Greater LDA topic-distribution divergence & Creativity \\
Phase alignment & Llama higher & Greater phase-vocabulary alignment & Spontaneous phase awareness \\
\bottomrule
\end{tabular}
\end{table}

\begin{table}[htbp]
\caption{Descriptive statistics for human PM turns and Llama 3.1 8B Instruct responses. Types are unique lowercased word tokens. Mean sentences count terminal punctuation.}
\label{tab:descriptive}
\centering
\small
\begin{tabular}{lrrrrr}
\toprule
\textbf{Source} & \textbf{N} & \textbf{Words} & \textbf{SD} & \textbf{Types} & \textbf{Sents} \\
\midrule
Human PM & 199 & 23.5 & 12.7 &  725 & 1.5 \\
LLM $T=0.3$ & 199 & 40.5 &  7.9 & 1031 & 1.8 \\
LLM $T=0.7$ & 199 & 38.5 &  7.5 & 1125 & 1.6 \\
LLM $T=1.0$ & 199 & 37.6 &  8.5 & 1196 & 1.5 \\
\bottomrule
\end{tabular}
\end{table}

LLM and human facilitators exhibit distinguishable behavioral patterns even when responding to the same conversational contexts. The primary family is these five tests only. The secondary phase-label family is corrected separately and is not part of this decision. The exploratory directiveness row in Table~\ref{tab:main_results} is not interpreted.

\begin{table}[htbp]
\caption{Paired Wilcoxon signed-rank tests: human PM turns versus Llama 3.1 8B Instruct ($T = 0.7$, $n = 199$ pairs). $p_{\text{Bonf}} = \min(1, 5p_{\text{raw}})$ for the five confirmatory dimensions and is compared with $0.05$ (equivalently, raw $p$ with $0.010$). Effect size $r = |Z|/\sqrt{n}$. The 95\% confidence interval is a bootstrap with 2{,}000 resamples.}
\label{tab:main_results}
\centering
\small
\begin{tabular}{@{}lcccccc@{}}
\toprule
\textbf{Dimension} & \textbf{H Mdn} & \textbf{L Mdn} & \textbf{$r$} & \textbf{95\% CI} & \textbf{$p_{\text{Bonf}}$} & \textbf{Sig.} \\
\midrule
Contextual distance & 0.742 & 0.593 & 0.644 & [0.56, 0.72] & $<$0.001 & *** \\
Lexical elaboration & 0.397 & 0.416 & 0.288 & [0.16, 0.40] & $<$0.001 & *** \\
Empathy/hedge density & 0.000 & 0.000 & 0.203 & [0.07, 0.34] & 0.021 & * \\
Topic divergence & 0.540 & 0.744 & 0.703 & [0.63, 0.76] & $<$0.001 & *** \\
Phase-vocabulary alignment$^{a}$ & 0.000 & 0.000 & 0.382 & [0.25, 0.50] & $<$0.001 & *** \\
\midrule
Directiveness (unvalidated)$^{b}$ & 0.021 & 0.036 & 0.026 & [0.00, 0.17] & 0.718 & ns \\
\bottomrule
\end{tabular}

\vspace{4pt}
{\footnotesize $^{a}$The primary prompt included the phase label. A secondary generation without it is in Table~\ref{tab:nophase}. $^{b}$Excluded from confirmatory inference: domain $\kappa = -0.02$. Stars use the adjusted $p$-value: one star means $p_{\text{Bonf}}<0.05$, and three stars mean $p_{\text{Bonf}}<0.001$. The empathy raw $p$-value is $0.00415$.}
\end{table}

\begin{figure}[htbp]
    \centering
    \includegraphics[width=\textwidth]{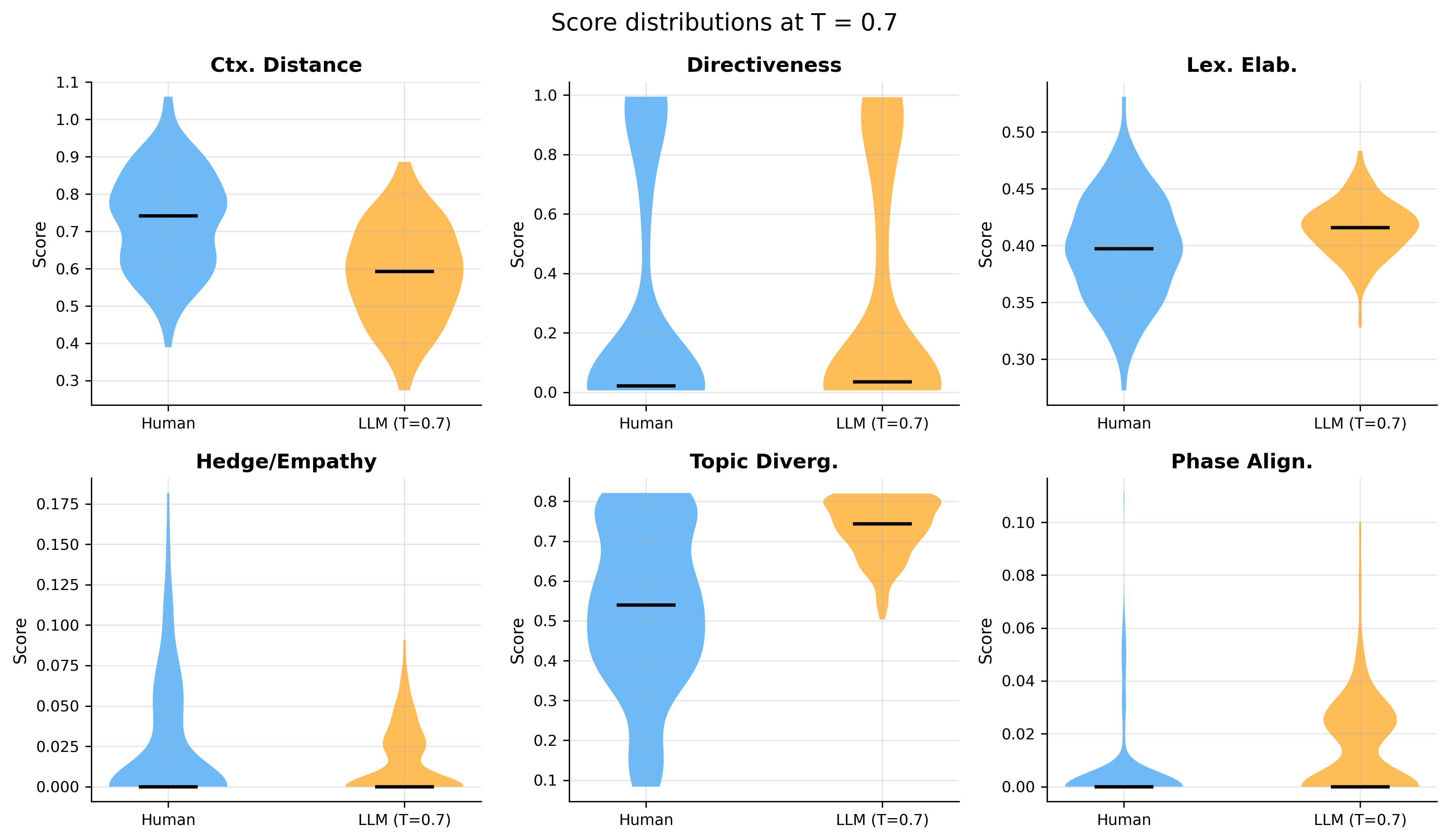}
    \caption{Score distributions for human facilitation and Llama~3.1 8B Instruct at $T = 0.7$. Horizontal lines are medians. Hedge/Empathy is lexical density, not a psychological rating. Directiveness is exploratory and unvalidated and is not a confirmatory dimension.}
    \label{fig:distributions}
\end{figure}

\subsection{Contextual semantic distance}
Human PM facilitation turns showed greater contextual distance from the preceding conversational context than Llama~3.1 8B Instruct responses in \texttt{all-MiniLM-L6-v2} embedding space ($r = 0.644$, 95\% CI [0.56, 0.72]). TF--IDF cosine distance did not reproduce the contrast ($r = 0.002$, $p = 0.978$). The human-higher contrast was observed with \texttt{all-MiniLM-L6-v2} but was not reproduced by the TF--IDF representation.

\subsection{Topic divergence}
Llama~3.1 8B Instruct responses showed greater divergence from the preceding context in LDA topic-distribution space than human PM facilitation turns ($r = 0.703$, 95\% CI [0.63, 0.76]). The same Llama-higher direction held when the topic model was refit at $K = 5, 10, 15, 20$, and $30$ (paired $r = 0.715$, $0.655$, $0.598$, $0.599$, and $0.445$). We therefore interpret the result as divergence in fitted topic distributions. It is a different quantity from the TF--IDF distance.

\subsection{Lexical elaboration proxy}
LLM-generated facilitation turns showed higher lexical elaboration than human facilitation turns ($r = 0.288$, 95\% CI [0.16, 0.40]). The paired difference is substantially associated with human word count (Spearman $\rho = -0.596$, $p < 0.001$). Because of that association, we treat the score as surface lexical elaboration.

\subsection{Empathy and hedging}
Human facilitation turns showed higher hedge-and-empathy lexical density than Llama-generated facilitation ($r = 0.203$, 95\% CI [0.07, 0.34]). The raw $p$-value is $0.00415$, below $0.010$. Equivalently, $p_\text{Bonf} = \min(1, 5p) = 0.021$, below $0.05$. Under that single five-test rule the contrast is significant. It is also the weakest confirmatory effect, and both medians are zero. Human turns contain 3 empathy-only dictionary hits and 116 hedge hits. Llama responses at $T = 0.7$ contain 30 empathy-only hits and 94 hedge hits. Because the signal is dominated by hedge terms, we interpret it as lexical hedging rather than psychological empathy.

\subsection{Phase-vocabulary alignment}
In the primary label-present generation, Llama responses showed greater phase-vocabulary alignment than human turns ($r = 0.382$, 95\% CI [0.25, 0.50]). Both medians are zero (means $0.0050$ and $0.0149$). The phase label was in the prompt, so the primary contrast is instruction-conditioned. Figure~\ref{fig:means} in the appendix prints the raw means on separate scales so the smaller empathy and phase means remain visible.

\subsection{Secondary phase-label sensitivity}
\label{sec:nophase}
\begin{figure}[htbp]
    \centering
    \includegraphics[width=0.72\textwidth]{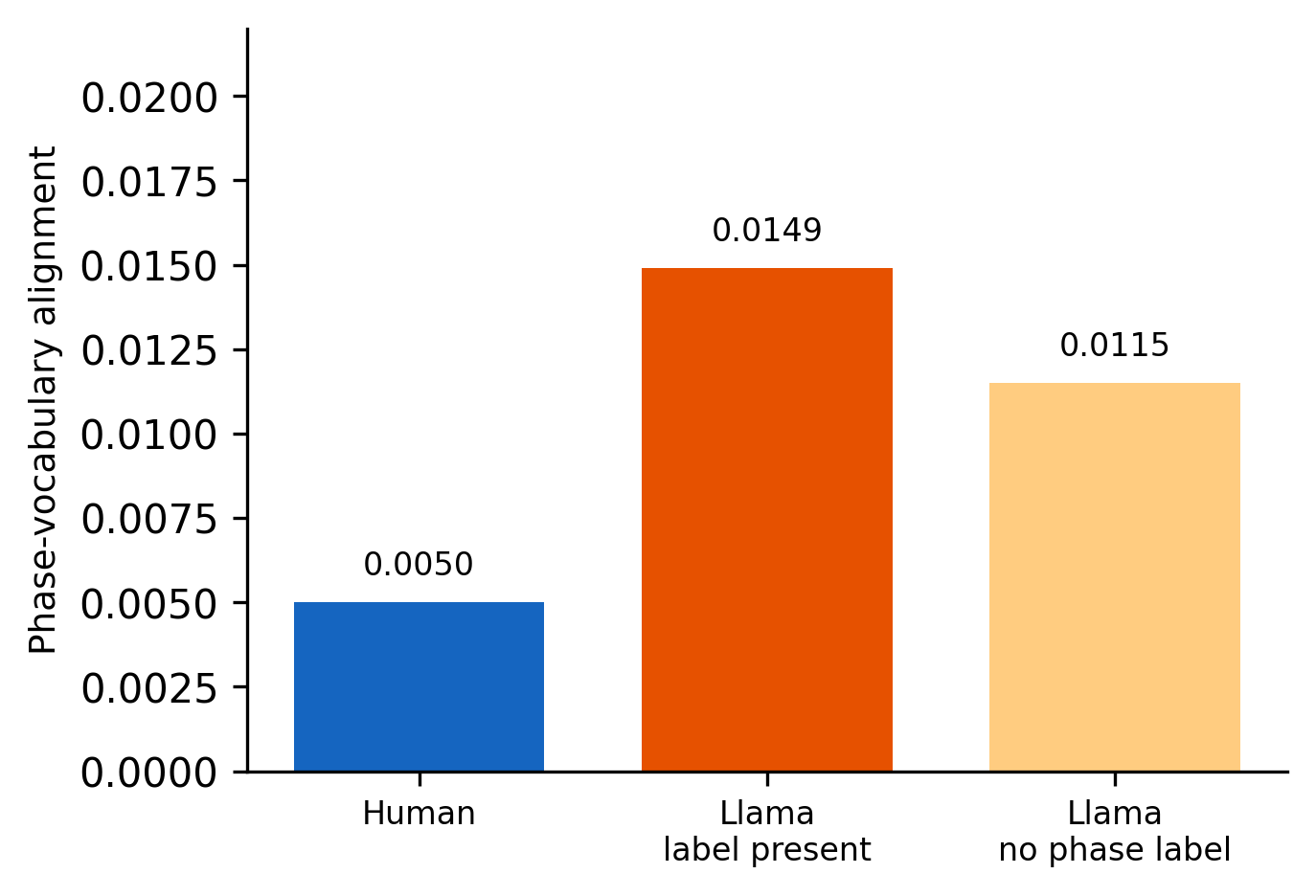}
    \caption{Phase-vocabulary means at $T = 0.7$. Removing the explicit phase label did not eliminate the Llama-higher contrast. The label-present versus no-label difference was not statistically significant after correction. The system prompt still requested a design-facilitation move.}
    \label{fig:nophase}
\end{figure}

The no-phase generation is a secondary check, not a sixth confirmatory dimension. Rescoring the stored label-present responses reproduces the published phase means (human $0.0050$; Llama $T = 0.7$ with the label $0.0149$). Without the label the $T = 0.7$ mean is $0.0115$ (Figure~\ref{fig:nophase}). All three medians are zero. Two tests at $T = 0.7$ share one Bonferroni correction, separate from the five-dimension family (Table~\ref{tab:nophase}).

Removing the explicit phase label did not eliminate the Llama-higher phase-vocabulary alignment contrast ($r = 0.294$, 95\% CI [0.16, 0.42], raw $p = 3.3 \times 10^{-5}$, secondary $p = 6.6 \times 10^{-5}$). The mean was somewhat lower without the label. The label-present versus no-label comparison was not statistically significant after correction ($r = 0.123$, 95\% CI [0.01, 0.26], raw $p = 0.082$, secondary $p = 0.163$), so we do not conclude that explicitly naming the phase increases the alignment score. The same label contrast is nonsignificant at $T = 0.3$ and $T = 1.0$ after correction across the three temperatures (raw $p = 0.042$ and $0.036$). The system prompt still requested a design-facilitation move, so this check does not show spontaneous phase awareness.

\begin{table}[htbp]
\caption{Secondary phase-label sensitivity at $T = 0.7$ ($n = 199$ matched contexts). The two raw $p$-values are Bonferroni-adjusted together. This family is separate from the five confirmatory dimensions. Effect size $r = |Z|/\sqrt{n}$, with the same 2{,}000-resample bootstrap as the primary tests.}
\label{tab:nophase}
\centering
\small
\begin{tabular}{@{}lccccc@{}}
\toprule
\textbf{Contrast} & \textbf{Means} & \textbf{$r$} & \textbf{95\% CI} & \textbf{Raw $p$} & \textbf{Sec.\ $p$} \\
\midrule
Human vs.\ no label & 0.0050 vs.\ 0.0115 & 0.294 & [0.16, 0.42] & $<$0.001 & $<$0.001 \\
Label vs.\ no label & 0.0149 vs.\ 0.0115 & 0.123 & [0.01, 0.26] & 0.082 & 0.163 \\
\bottomrule
\end{tabular}
\end{table}

\subsection{Robustness and temperature}
The checks below answer alternative explanations of the observed differences. They are not a second primary family.

\begin{table}[htbp]
\caption{Friedman tests of scores across $T \in \{0.3, 0.7, 1.0\}$ within the same 199 contexts. Bonferroni correction uses the five confirmatory dimensions.}
\label{tab:temp}
\centering
\small
\begin{tabular}{@{}lcccc@{}}
\toprule
\textbf{Dimension} & \textbf{$\chi^2$} & \textbf{$p$} & \textbf{$p_{\text{Bonf}}$} & \textbf{Sig.} \\
\midrule
Contextual distance & 0.44 & 0.802 & 1.000 & --- \\
Lexical elaboration & 4.04 & 0.133 & 0.664 & --- \\
Empathy/hedge density & 2.21 & 0.332 & 1.000 & --- \\
Topic divergence & 5.46 & 0.065 & 0.327 & --- \\
Phase-vocabulary alignment & 5.61 & 0.060 & 0.302 & --- \\
\midrule
Directiveness (exploratory) & 0.85 & 0.654 & 0.654 & --- \\
\bottomrule
\end{tabular}
\end{table}

Within-context Friedman tests found no statistically detectable temperature effect on the confirmatory dimensions across $T \in \{0.3, 0.7, 1.0\}$ after Bonferroni correction (Table~\ref{tab:temp}). Descriptive temperature correlations are in Appendix~\ref{app:robustness}. Those correlations stack repeated contexts, so the Friedman tests remain the inferential result.

Length-matched pairs ($n = 73$) kept the same direction for contextual distance ($r = 0.514$), lexical elaboration ($r = 0.807$), topic divergence ($r = 0.712$), and phase alignment ($r = 0.487$). Response length substantially accounts for the elaboration contrast, which is why that proxy stays a surface measure. Empathy-and-hedge density did not meet the primary raw threshold in the length-matched subset ($r = 0.231$, $p = 0.048$). Removing 63 verbatim duplicate contexts leaves 136 pairs. The five confirmatory directions remain and still meet the same Bonferroni rule (contextual distance $r = 0.678$, lexical elaboration $r = 0.326$, empathy-and-hedge $r = 0.276$, raw $p = 0.001$, $p_\text{Bonf} = 0.006$, topic divergence $r = 0.712$, phase alignment $r = 0.318$). Repetition does not by itself produce the directional pattern. Topic divergence stays Llama-higher from $K = 5$ to $K = 30$, so the direction is not an artifact of $K = 15$ alone. TF--IDF does not reproduce the embedding-distance result, so the contrast should not be treated as representation-independent semantic distance. Mixed-effects estimates have the same sign as the paired differences on all five confirmatory dimensions (Appendix~\ref{app:robustness}).

\section{Discussion}
\label{sec:discussion}
\subsection{A behavioral profile of LLM facilitation}
Context-matched LLM and human facilitation turns exhibit a distinguishable behavioral profile even when responding to the same conversational contexts. The observed differences concern properties of the facilitation language. These differences should not be interpreted as a ranking of facilitation quality.

\subsection{Contextual alignment and divergence}
Human PM facilitation turns exhibited greater contextual distance from the preceding conversational context than Llama~3.1 8B Instruct responses in \texttt{all-MiniLM-L6-v2} embedding space ($r = 0.644$). Llama~3.1 8B Instruct responses exhibited greater divergence from the preceding context in LDA topic-distribution space, with the direction remaining stable across topic counts from 5 to 30. The two contrasts are not interchangeable: TF--IDF cosine distance did not reproduce the embedding result ($r = 0.002$, $p = 0.978$). For human--AI interaction, the practical point is that distance from context depends on the representation. For computational social science, the same paired design can carry more than one distributional measure without treating them as one construct.

\subsection{Surface lexical behavior}
The higher lexical-elaboration score for LLM turns moves with response length (Spearman $\rho = -0.596$, $p < 0.001$). Length matching changed the size of the effect ($r = 0.807$ at $n = 73$) and left the direction unchanged. Because the proxy varies with response length, comparisons across prompting conditions should account for this confound.

\subsection{Human hedge/empathy language}
The human advantage on hedge-and-empathy density ($r = 0.203$, raw $p = 0.00415$, $p_\text{Bonf} = 0.021$) reflects primarily the greater contribution of hedge terms to the dictionary-based score, not psychological empathy. Both medians are zero. In the length-matched subset the contrast did not meet the primary raw threshold ($p = 0.048$).

\subsection{Phase-conditioned facilitation language}
Removing the phase label left the Llama-higher contrast in place ($r = 0.294$). The label-present versus no-label comparison was not statistically significant after correction. We therefore do not conclude that explicitly naming the phase increases the alignment score. The system prompt still requested a design-facilitation move. Full-session lengths are not stored, so other temporal boundaries were not re-estimated. The evaluated set contains 149 Ideate turns and 50 Prototype/Evaluate turns.

\subsection{Implications for future LLM-facilitation research}
The paired design offers a comparison point for later facilitation studies. Extending the analysis to other models, prompts, domains, or meeting structures would require independent evaluation. The present comparison also sits alongside, rather than in place of, work on facilitation systems \citep{dong2024facilitation}, intervention timing \citep{tsirmpas2026facilitate}, and group outcomes \citep{alsobay2026table}. For HCI, the measured differences are evidence about human--AI facilitation behavior, limited to measurable language differences. For NLP and computational social science, the comparison illustrates how conclusions about conversational behavior can depend on the representation used, as seen across embedding, TF--IDF, and LDA measures. For design research, it characterizes facilitation language in collaborative design meetings. For LLM research, it describes the facilitation-language differences observed when an LLM is assigned a facilitation role. That role assignment is not recognition of design phases.

Directiveness remains unresolved ($\kappa = -0.0231$). No conclusion about human versus LLM directiveness is warranted.

Several boundaries remain. The model is Llama~3.1 8B Instruct in 4-bit. The human turns come from project managers in the AMI remote-control scenario, rather than a sample of professional facilitators across workplace settings. Phase bins are author-defined. Lexical elaboration is length-confounded, and the empathy lexicon is hedge-dominated. The embedding contrast is not reproduced by TF--IDF. Topic divergence uses a heuristically chosen 15-topic model, although the direction is stable at other values of $K$. Sixty-three contexts are duplicated, and the directional pattern is still present in the 136 unique pairs. The paired test does not by itself model session clustering; the mixed-effects check only partly addresses that.

\section{Conclusion}
We report a context-matched behavioral profile of LLM and human facilitation in collaborative design meetings. The sampled human Project Manager turns are higher on embedding distance and on hedge-and-empathy lexical density. Llama~3.1 8B Instruct is higher on topic divergence, surface lexical elaboration, and instruction-conditioned phase-vocabulary alignment. The no-phase check leaves the Llama-higher phase contrast in place, without showing spontaneous phase awareness. The findings concern measurable properties of facilitation language. They are not ratings of facilitation quality, creativity, psychological empathy, or general intelligence. These results provide a reference point for future research comparing LLM facilitators across models, prompts, domains, and collaborative settings.

\section*{Data Availability}
The AMI Meeting Corpus is distributed by its maintainers at \url{https://groups.inf.ed.ac.uk/ami/corpus/}. We do not redistribute that corpus. The analysis code, derived feature tables, and generated facilitation responses used in the comparison are available at \url{https://github.com/Ayaan577/llm-facilitation-behavior}.

\FloatBarrier
\bibliographystyle{unsrtnat}
\bibliography{references}

\appendix
\FloatBarrier
\section{Additional displays and robustness}
\label{app:robustness}
\begin{figure}[htbp]
    \centering
    \includegraphics[width=0.78\textwidth]{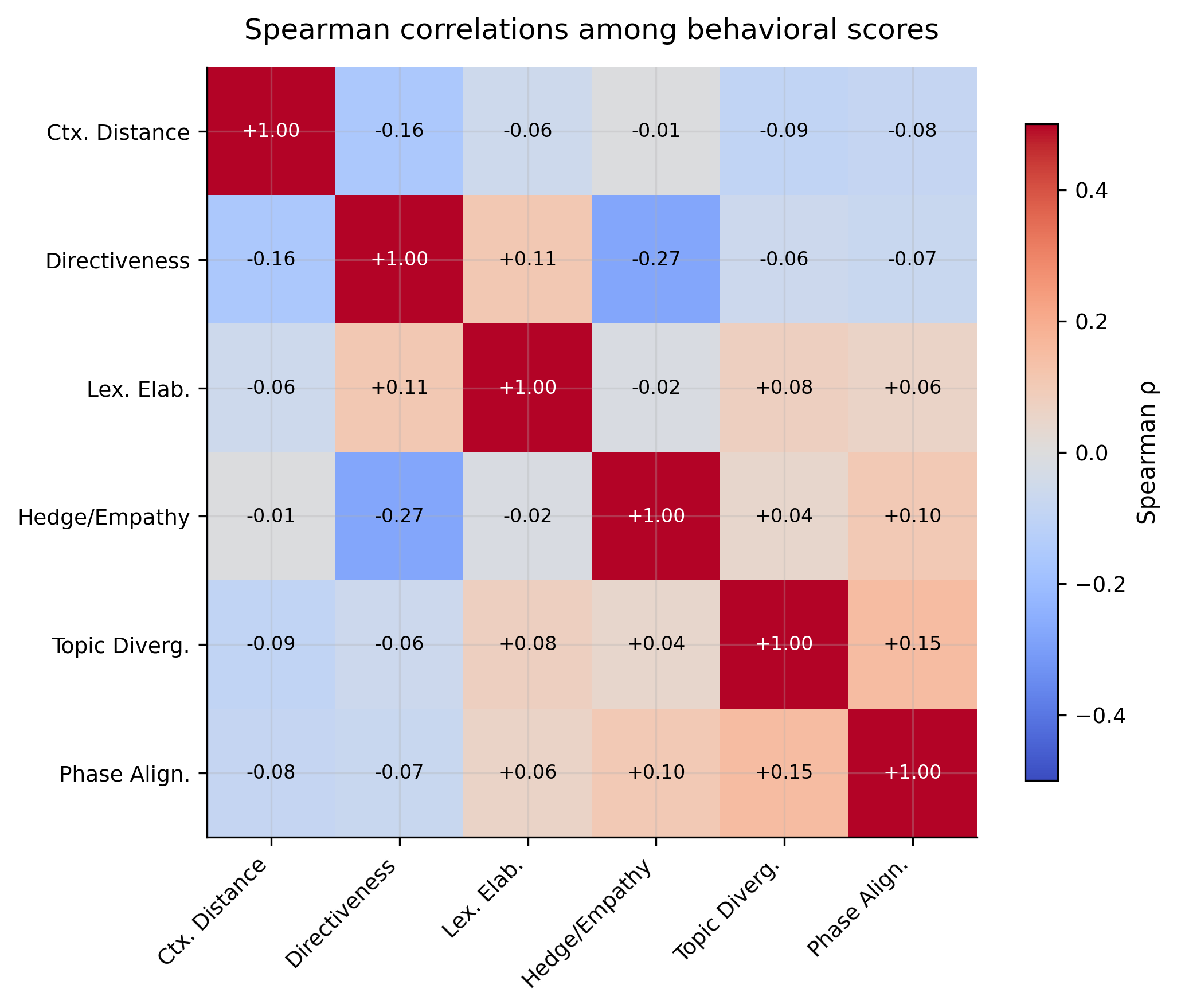}
    \caption{Spearman correlations among the six scores ($N = 796$), including the unvalidated directiveness score.}
    \label{fig:corrmatrix}
\end{figure}

\begin{figure}[htbp]
    \centering
    \includegraphics[width=0.86\textwidth]{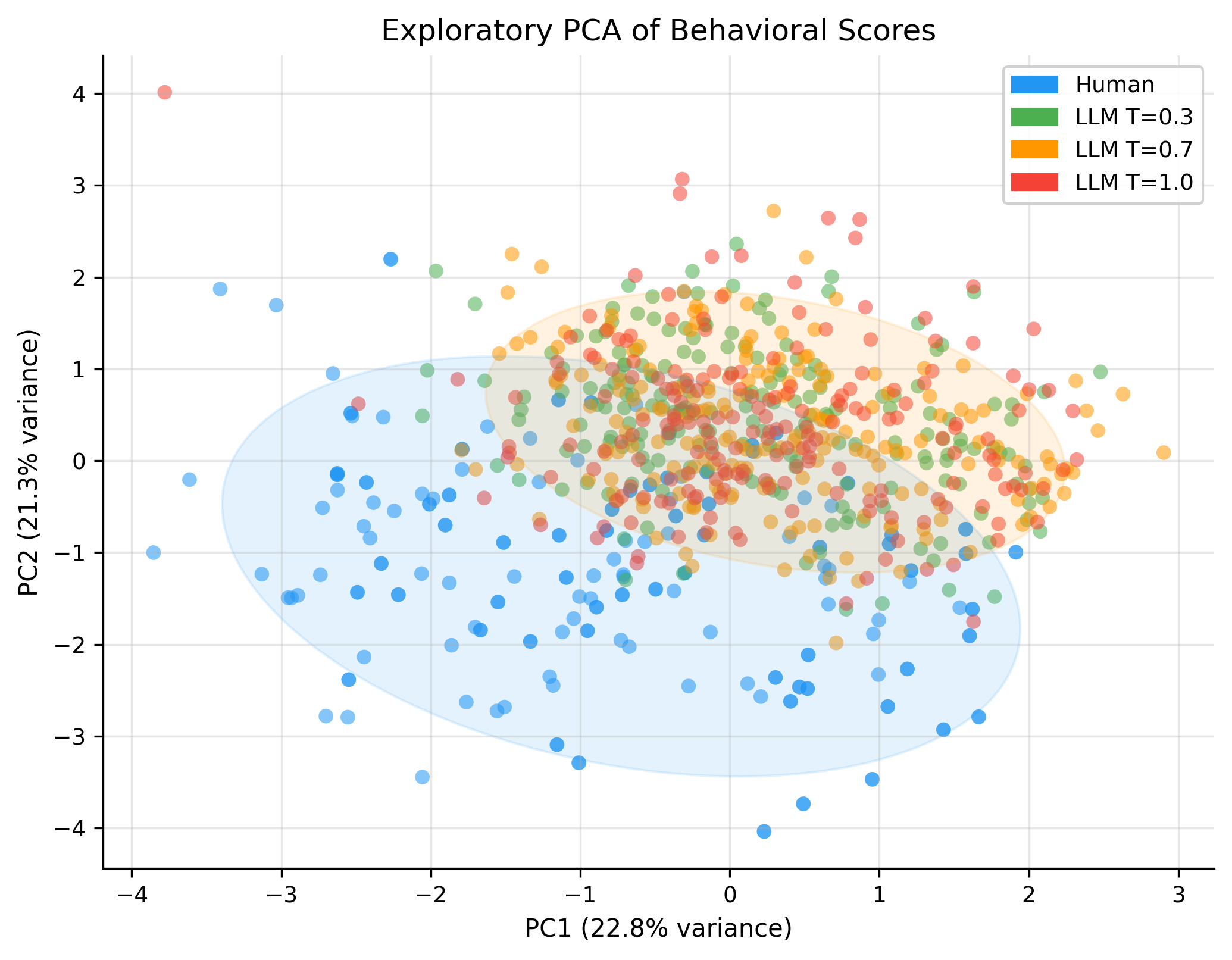}
    \caption{Exploratory PCA of behavioral scores for human turns and Llama responses at all three temperatures, fit on all 796 turns. The axes are principal components of the score matrix, not a validated style space.}
    \label{fig:pca}
\end{figure}

\begin{figure}[htbp]
    \centering
    \includegraphics[width=\textwidth]{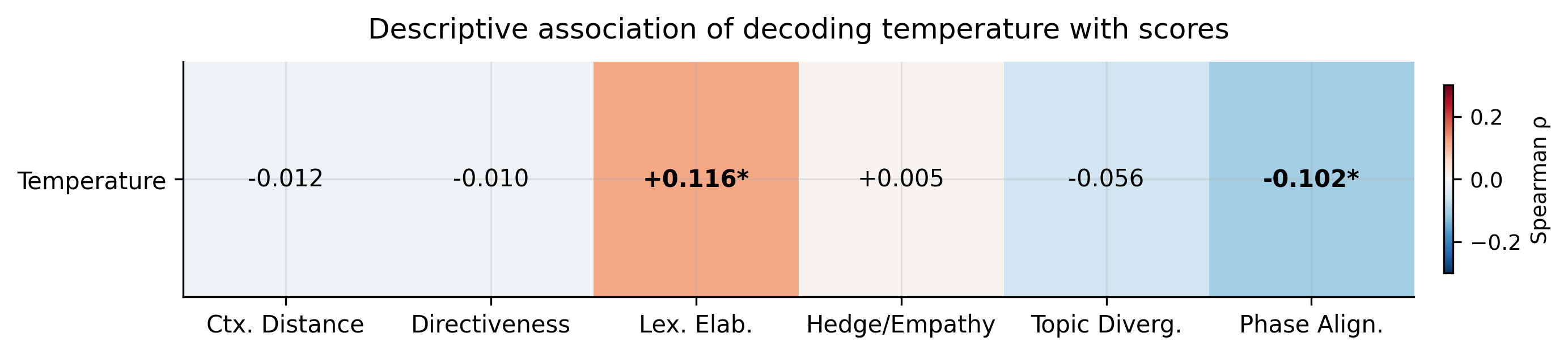}
    \caption{Descriptive Spearman correlations between decoding temperature and scores on 597 stacked responses. These are not the within-context Friedman tests.}
    \label{fig:heatmap}
\end{figure}

Pairwise Spearman correlations among the six scores remain weak ($|\rho| < 0.28$; Figure~\ref{fig:corrmatrix}). An exploratory principal component analysis of all 796 turns attributes 22.8\% of variance to PC1 and 21.3\% to PC2 (44.1\% together; Figure~\ref{fig:pca}). Human turns lie toward negative PC1 and LLM responses toward positive PC1. This decomposition is descriptive. It is not a validated facilitation-style construct. Figure~\ref{fig:heatmap} shows descriptive associations with temperature. The inferential temperature tests are the Friedman tests in Table~\ref{tab:temp}.

\begin{figure}[htbp]
    \centering
    \includegraphics[width=\textwidth]{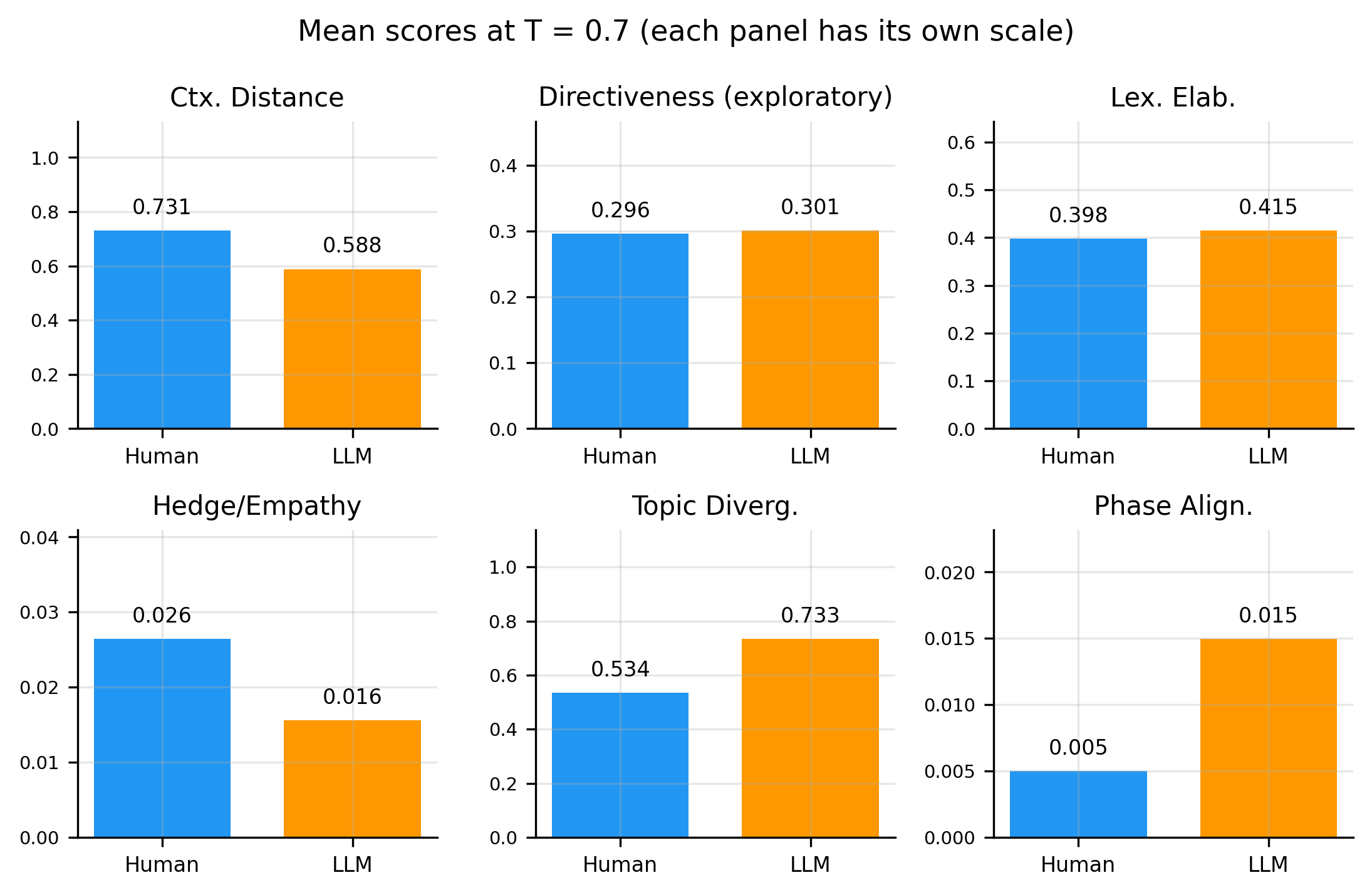}
    \caption{Mean scores for human facilitation and Llama responses at $T = 0.7$. Each panel starts at zero and uses its own scale. Printed numbers are the raw means. Hedge/Empathy is lexical density. Directiveness is exploratory and unvalidated and is not a confirmatory dimension.}
    \label{fig:means}
\end{figure}

\begin{table}[htbp]
\caption{Robustness checks: length-matched pairs ($n = 73$), Spearman correlation of the paired difference with human word count, and a mixed-effects estimate with a random intercept for context. Primary $r$ is from the paired Wilcoxon test ($n = 199$).}
\label{tab:robustness}
\centering
\small
\begin{tabular}{@{}lcccccc@{}}
\toprule
\textbf{Dimension} & \textbf{Prim.\ $r$} & \textbf{Len.\ $r$} & \textbf{Part.\ $\rho$} & \textbf{ME $\beta$} & \textbf{$p_{\text{Bonf}}$} & \textbf{Note} \\
\midrule
Contextual distance & 0.644 & 0.514 & $-$0.137 & $-$0.143 & $<$0.001 & Stable \\
Topic divergence & 0.703 & 0.712 & $-$0.027 & +0.199 & $<$0.001 & Stable \\
Lexical elaboration & 0.288 & 0.807 & $-$0.596 & +0.017 & $<$0.001 & Length-confounded \\
Phase alignment$^{a}$ & 0.382 & 0.487 & $-$0.059 & +0.010 & $<$0.001 & Instruction-conditioned \\
Empathy/hedge density & 0.203 & 0.231 & +0.068 & $-$0.011 & 0.021 & Hedge proxy \\
Directiveness (exploratory) & 0.026 & --- & --- & +0.005 & 0.718 & Unvalidated \\
\bottomrule
\end{tabular}

\vspace{4pt}
{\footnotesize $^{a}$The primary prompt included the phase label. The separate no-label comparison did not show a significant label effect after correction.}
\end{table}

\end{document}